\documentclass[lettersize,journal]{IEEEtran}
\usepackage{amsmath,amsfonts}
\usepackage{algorithmic}
\usepackage{algorithm}
\usepackage{array}
\usepackage[caption=false,font=normalsize,labelfont=sf,textfont=sf]{subfig}
\usepackage{textcomp}
\usepackage{stfloats}
\usepackage{url}
\usepackage{multirow}
\usepackage{booktabs}
\usepackage{indentfirst} 
\usepackage{verbatim}
\usepackage{enumitem}
\usepackage{graphicx}
\usepackage{cite}
\usepackage{hyperref}
\usepackage{cleveref}
\usepackage[normalem]{ulem}
\usepackage{xspace}
\usepackage[T1]{fontenc}
\useunder{\uline}{\ul}{} 
\begin{document}

\title{High-fidelity and Lip-synced Talking Face Synthesis via Landmark-based Diffusion Model}

\author{
        Weizhi~Zhong, 
        Junfan~Lin, 
        Peixin~Chen, 
        Liang~Lin, 
        and Guanbin~Li
\thanks{This work was supported in part by the National Natural Science Foundation of China (NO.~62322608). (Corresponding author is Guanbin Li).}
\thanks{W.~Zhong, J.~Lin, P.~Chen, L.~Lin and G.~Li are with the School of Computer Science and Engineering, Sun Yat-sen University, Guangzhou, China. 
E-mail: zhongwzh5@mail2.sysu.edu.cn, linjf8@mail2.sysu.edu.cn, chenpx28@mail2.sysu.edu.cn, linliang@ieee.org, liguanbin@mail.sysu.edu.cn}

}


\markboth{Journal of \LaTeX\ Class Files,~Vol.~14, No.~8, May~2024}%
{Shell \MakeLowercase{\textit{et al.}}: A Sample Article Using IEEEtran.cls for IEEE Journals}

\IEEEpubid{0000--0000/00\$00.00~\copyright~2024 IEEE}

\maketitle

\begin{abstract}
Audio-driven talking face video generation has attracted increasing attention due to its huge industrial potential.
Some previous methods focus on learning a direct mapping from audio to visual content. Despite progress, they often struggle with the ambiguity of the mapping process, leading to flawed results. 
An alternative strategy involves facial structural representations (e.g., facial landmarks) as intermediaries. This multi-stage approach better preserves the appearance details but suffers from error accumulation due to the independent optimization of different stages.
Moreover, most previous methods rely on generative adversarial networks, prone to training instability and mode collapse.
To address these challenges, our study proposes a novel landmark-based diffusion model for talking face generation, which leverages facial landmarks as intermediate representations while enabling end-to-end optimization. 
Specifically, we first establish the less ambiguous mapping from audio 
to landmark motion of lip and jaw. 
Then, we introduce an innovative conditioning module called TalkFormer to align the synthesized motion with the motion represented by landmarks via differentiable cross-attention, which enables end-to-end optimization for improved lip synchronization. 
Besides, TalkFormer employs implicit feature warping to align the reference image features with the target motion for preserving more appearance details.
Extensive experiments demonstrate that our approach can synthesize high-fidelity 
 and lip-synced talking face videos, preserving more subject appearance details from the reference image.
\end{abstract}


\section{Introduction}
\label{sec:intro}
\IEEEPARstart{A}{udio}-driven talking face video generation, a challenging task of cross-modal synthesis, aims to create talking videos with lip movements that are accurately synchronized with the input audio. This task has attracted increasing interest from the research community due to its broad applications such as visual dubbing~\cite{xie2021towards}, virtual avatars~\cite{zhou2022dialoguenerf}, and digital humans~\cite{thies2020neural}.
To generate high-fidelity talking face videos, a prevalent manner is to gather data of the target individual to learn a person-specific model~\cite{guo2021ad,huang2023parametric,du2023dae,ye2022geneface}. While effective, these person-specific methods are hindered by the costly data collection and extensive training process.
In contrast, person-generic methods can generalize to unseen subjects without further training. However, these methods often grapple with challenges in maintaining the appearance details of subjects as well as lip-audio synchronization. In this study, we endeavor to develop a person-generic talking face generation framework to generate high-fidelity facial details for general subjects while ensuring accurate lip-audio synchronization.

\IEEEpubidadjcol

\begin{figure}[t]
  \centering 
  \includegraphics[width=\columnwidth]{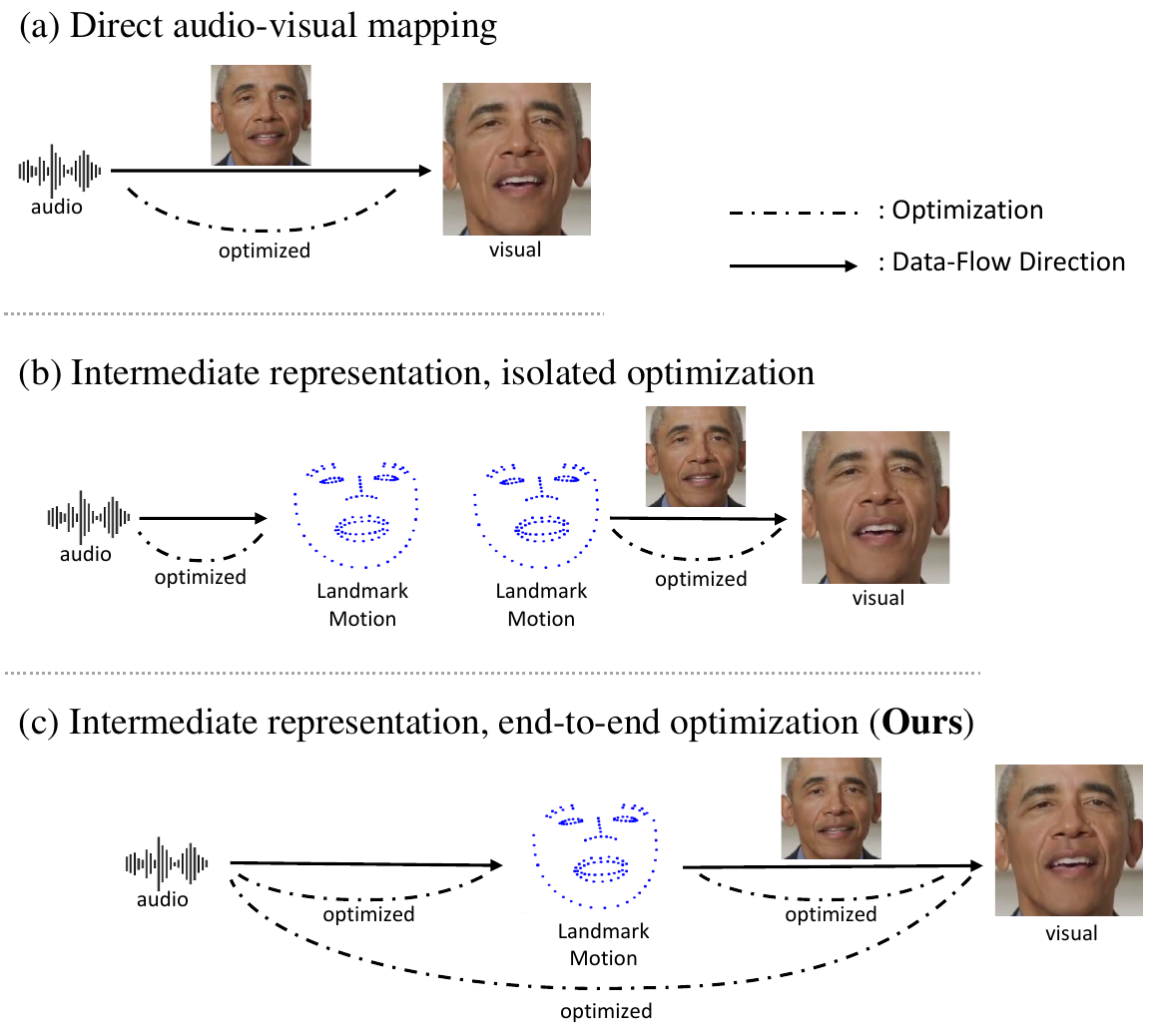}
  \caption{An illustration of three strategies to learn the audio-visual relationship for talking face generation. \textbf{(a)}~\textbf{Previous} methods optimize a network to learn the direct audio-visual mapping, resulting in flawed results. 
  \textbf{(b)}~\textbf{Previous} methods optimize a network to learn the mapping from audio to landmark motion, then separately optimize another mapping from the motion representation to realistic face, suffering from the inaccuracies of pre-estimated intermediate representation. 
  \textbf{(c)}~\textbf{Our} method leverages facial landmarks as intermediate representation while enabling end-to-end optimization to reduce the errors accumulation resulting from pre-estimated landmark inaccuracies.   
  } 
  \label{fig:Intro_demo} 
\end{figure}

To achieve faithful talking face generation, two critical issues need to be considered: the high fidelity of the subject appearance details and the synchronization of lip movement with the audio input.
To generate lip-synced talking videos, prior methods~\cite{prajwal2020lip,shen2023difftalk,wang2023progressive} directly model the mapping between audio signal and visual content. However, such audio-visual mapping is often uncertain and ambiguous, as one phonetic unit potentially matches various visual forms due to the diversities in illumination, emotion, and appearance. This often leads to flawed results and the loss of details in subject appearance.
To ease the ambiguity for improved fidelity of subject 
 appearance, another line of works~\cite{zhong2023identity,xie2021towards,zhou2020makelttalk,gururani2023space,zhang2021flow,sinha2022emotion} instead first establish a correlation between the audio signals and intermediate structural representation such as facial landmarks or 3D Morphable Model (3DMM)~\cite{blanz1999morphable}, which primarily reflect the motion information and facilitate the less ambiguous mapping from audio to motion. 
Then, another stage converts these motion representations into realistic facial images.
However, a notable drawback of these approaches is the isolated training of different stages, potentially resulting in inaccuracies in lip-audio synchronization stemming from error in the pre-estimated structural representations. 
Additionally, prior methods\cite{prajwal2020lip,shen2023difftalk,sinha2022emotion,zhong2023identity} frequently rely on either misaligned reference images or images warped by imprecise optical flow predictions as generation conditions, neglecting the influence of such misalignment on the preservation of appearance details. 
Moreover, most previous approaches employ generative adversarial networks (GANs)~\cite{goodfellow2020generative} for talking face generation, which often suffers from training instability and the issue of mode collapse.

To tackle the challenges above, we propose an innovative landmark-based diffusion model to learn the audio-visual relation. 
\Cref{fig:Intro_demo} compares previous methods and ours. 
Our method utilizes facial landmarks as intermediate representation to ease the ambiguity of audio-visual mapping while enabling the end-to-end optimization of distinct stages, facilitating the generation of high-fidelity and lip-synced talking face video.      
Specifically, our two-stage approach initially converts the input audio signal into a set of landmark representations using a landmark completion module~\cite{zhong2023identity}. 
To enable the end-to-end optimization of distinct stages and align the synthesized motion with the motion represented by landmarks, we devise a novel conditioning module called TalkFormer to integrate landmark representations into the diffusion model using differentiable cross-attention. This end-to-end approach significantly reduces error accumulation resulting from pre-estimated landmark inaccuracies, thereby improving lip-audio synchronization. 
Besides, to enhance the fidelity of facial appearance, our approach converts a reference facial image into multi-scale features capturing intricate details. 
Then, our proposed TalkFormer module spatially aligns these features with the target motion using an implicit warping technique. 
Without using imprecise optical flow, the implicit warping automatically establishes semantic correlations for improved alignment between the reference features and the synthesized content. 
These aligned reference features are then integrated into the denoising process, enhancing the preservation of subject appearance details from reference image. 
Our extensive experiments validate our framework's effectiveness in producing realistic talking face videos with high-fidelity subject appearance and lip movements accurately synchronized to the input audio.
We summarize our contributions as follows:
\begin{itemize}
     \item We propose a novel method to learn the audio-visual relationship for generating high-fidelity and lip-synced talking face videos, utilizing facial landmarks as intermediate representation to ease the ambiguity of audio-visual mapping while enabling end-to-end optimization to minimize error accumulation.
    \item  We introduce a novel conditioning module, TalkFormer, to align the synthesized motion with the motion represented by landmarks in a differentiable manner, enabling the joint optimization of distinct stages.
    Additionally, TalkFormer aligns the reference image features with the target motion based on semantic correlations, enhancing the preservation of subject appearance details.
    \item  We conduct comprehensive experiments to demonstrate our method's effectiveness in producing high-fidelity and lip-synced talking face videos, which can generalize to any unseen subject without additional fine-tuning. 
\end{itemize}

\section{Related Work}

\subsection{Audio-Driven Talking Face Generation}
Audio-driven talking face video generation techniques can be mainly divided into two types: person-specific or person-generic.
Many person-specific methods can generate vivid videos~\cite{guo2021ad,huang2023parametric,du2023dae,ye2022geneface,shen2022learning,zhang2021facial}, but they require videos of the target subject for additional training. 
On the contrary, person-generic methods~\cite{prajwal2020lip,shen2023difftalk,zhong2023identity,zhou2021pose,xie2021towards,wang2023progressive,zhou2020makelttalk,zhang2021flow,gururani2023space} enable inference on unseen subjects without any retraining or fine-tuning. However, there is still a gap in achieving high-fidelity and lip-synced talking face generation for person-generic methods. 

Wav2Lip~\cite{prajwal2020lip}, PD-FGC~\cite{wang2023progressive}, and PC-AVS~\cite{zhou2021pose} attempt to generate lip-synced videos by directly conditioning the generator on audio representation. Nevertheless, these approaches exhibit notable flaws and loss of appearance details in the subjects due to the inherent uncertainty and ambiguity in audio-visual mapping.
IP-LAP~\cite{zhong2023identity} and other methods~\cite{xie2021towards,zhou2020makelttalk} propose a two-stage framework that utilizes facial landmarks as intermediate representation. However, the projection of their intermediate landmark representation into the sketch image is indifferentiable. Subsequently, an image-to-image translation network is employed to synthesize realistic faces from the sketch images. Therefore, these methods train distinct stages independently and suffer from the inaccuracies of pre-estimated landmarks. 
Besides, prior methods~\cite{zhang2021flow,sinha2022emotion,gururani2023space,zhang2023sadtalker} including IP-LAP~\cite{zhong2023identity} utilize estimated optical flow to align the reference image with target facial expression and pose, such that more appearance details from reference image can be preserved during the generation process. However, accurate optical flow estimation is challenging especially when there is significant variation in head pose, leading to distorted results.
To tackle the drawbacks of GANs~\cite{goodfellow2020generative} in unstable training and mode collapse, DiffTalk~\cite{shen2023difftalk} crafts a diffusion model to learn the direct audio-to-visual mapping for generalized talking face generation. It directly models the audio-to-lip translation with the landmarks of upper-half face concatenated as auxiliary condition. However, the landmarks of upper-half face are insufficient to alleviate the uncertainty of direct audio-to-visual mapping. 
Besides, its usage of the misaligned reference image hinders the preservation of subject appearance details from reference image. 
Recently, GAIA~\cite{he2023gaia} proposes to disentangle motion and appearance using VAE\cite{kingma2013auto} and utilizes diffusion models to predict motion from the speech. However, it can not achieve end-to-end learning of the framework to reduce error accumulation.
Besides, it leverages misaligned reference appearance features as generation conditions, which hinders the preservation of facial details from reference images.
%

Contrasting with prior approaches, our framework employs facial landmarks as intermediate representation while enabling end-to-end optimization. Our novel conditioning module TalkFormer integrates landmarks representation into diffusion models in a differentiable way and aligns the reference appearance features based on semantic correlations, facilitating the generation of high-fidelity and lip-synced talking face video.

\subsection{Diffusion Models}

Diffusion models~\cite{sohl2015deep,ho2020denoising} have recently emerged as a promising type of generative models, exhibiting superior generation power and enhanced training stability when compared to GANs~\cite{goodfellow2020generative}.
Denoising diffusion probabilistic models~(DDPM)~\cite{ho2020denoising} is a class of latent variable models that uses a diffusion process to add noise to the data gradually and learns a denoiser network to reverse the diffusion process. 
Recently, Denoising Diffusion Implicit Models~(DDIM)~\cite{song2020denoising} are proposed to accelerate the sampling process of diffusion models through a class of non-Markovian diffusion processes. Latent diffusion models~(LDM)~\cite{rombach2022high} apply diffusion model in the latent space of powerful pre-trained autoencoders to save computational resources while retaining the generation quality.
Diffusion models have recently demonstrated remarkable success in various synthesis tasks~\cite{kim2023diffusion,xing2023diffsketcher,zhao2023towards,wu2023tune,seo2023midms}, 
including text-to-image generation~\cite{zhang2023adding,mou2023t2i,rombach2022high}, image editing~\cite{hertz2022prompt,yang2023paint}, person image synthesis~\cite{bhunia2023person,shen2023advancing}, face video editing\cite{kim2023diffusion}, and face restoration\cite{zhao2023towards,yue2022difface}. 
Stable Diffusion~\cite{rombach2022high} conditions the latent diffusion model on the CLIP~\cite{radford2021learning} text embedding, achieving compelling text-to-image generation. 
ControlNet~\cite{zhang2023adding} proposes a neural network architecture that incorporates spatial conditions (e.g., sketch image) into pre-trained diffusion models.

For talking face video generation with diffusion models, previous methods\cite{shen2023difftalk,stypulkowski2024diffused,mukhopadhyay2024diff2lip,bigioi2024speech} make an early attempt to learn the direct audio-visual mapping by conditioning the diffusion model on audio feature, but often produce flawed results due to the ambiguity of mapping. 
To alleviate this problem, a straightforward method is to involve facial landmarks as intermediate representation and integrate the predicted landmarks from audio into the diffusion model using ControlNet~\cite{zhang2023adding} architecture.
However, a notable drawback of this manner is the isolated training of audio-to-landmark and landmark-to-video stages, resulting in sub-optimal lip synchronization. This is because the projection from landmarks coordinates into sketch image is indifferentiable.  
In contrast, in this study, we propose an innovative talking face video generation framework based on efficient latent diffusion models, which leverages facial landmarks as intermediate representation and enables end-to-end optimization of distinct stages. 
Our method first predict the lip and jaw landmarks coordinates from audio signals, and a novel conditioning module TalkFormer integrates the landmarks into diffusion model in a differentiable manner.

\section{Methodology}

\begin{figure*}[ht]  
  \centering 
  \includegraphics[width=\linewidth]{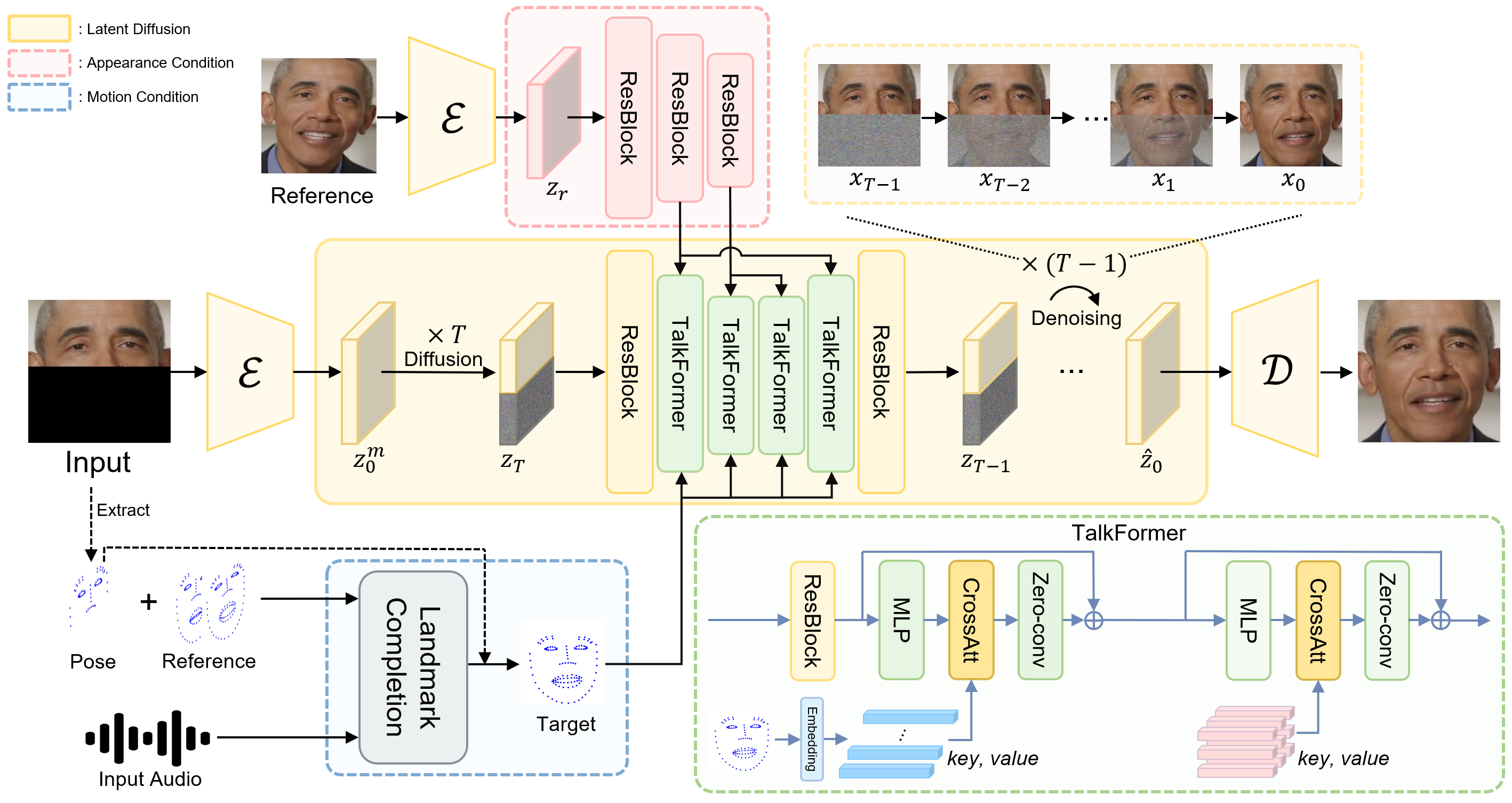}
  \caption{An overview of the proposed framework. The diffusion and reverse denoising operations are executed in the encoded latent space of an autoencoder $\mathcal{D}(\mathcal{E}(\cdot))$. 
  (1). Initially, the audio signal drives the completion of lip and jaw landmarks, guided by reference full-face and upper half-face pose landmarks. The completed lip and jaw landmarks are then combined with the input pose landmarks to form the target full-face landmarks.  
  (2). The conditioning module, TalkFormer, aligns the synthesized motion with the motion represented by target landmarks via differentiable cross-attention layers.
  To capture the intricate appearance details, a reference face image is encoded into multi-scale reference features. TalkFormer then aligns these features with the target motion via an implicit warping mechanism implemented by cross-attention layers. The skip-connections of U-Net are omitted for clarity.}
  \label{fig:framework}
\end{figure*}

The framework overview of our method is presented in \Cref{fig:framework}. For talking face video generation, our method inputs audio and a template video, masking the lower-half face.
The framework then inpaints these areas with realistic content synchronized with the audio. Information about the subject appearance and facial contours is derived from a single reference image and reference full-face landmarks from the template video, respectively.
More specifically, the audio signal drives the completion of lip and jaw landmarks, guided by reference full-face landmarks and pose landmarks detected from the upper-half face of the template video. The completed landmarks, as well as the reference image, are then fed into the latent diffusion model via TalkFormer, influencing the synthesized motion and appearance. 
During training, the network diffuses the lower half of the ground-truth face in latent space, focusing on noise reduction. 
Upcoming sections \Cref{pre_LDM}, \Cref{auio-landmark}, and \Cref{conteol_LDM} will introduce latent diffusion models and provide a comprehensive explanation of our approach.

\subsection{Preliminaries of Latent Diffusion Models}
\label{pre_LDM}
Latent diffusion models~\cite{rombach2022high} carry out diffusion and the denoising process in the encoded latent space of an autoencoder $\mathcal{D}(\mathcal{E}(\cdot))$, with $\mathcal{E}(\cdot)$ being the encoder and $\mathcal{D}(\cdot)$ being the decoder. A U-Net-based~\cite{ronneberger2015u} denoising network $\epsilon_{\theta}\left(z_{t}, t\right)$ is trained to predict the noise added to the image latent $z_{0}$, where $z_{0}=\mathcal{E}(x)$,  $x$ is the input image, $z_{t}$ represents the noisy version of $z_{0}$ at time step condition $t\in \{1,2,..., T\}$ and $\theta$ refers to the learnable parameters.  The optimization objective during training is as follows: 
\begin{equation}
\label{loss_ldm}
\mathcal{L}_{ldm}=\mathbb{E}_{z, \epsilon \sim \mathcal{N}(0,1), t}\left[\left\|\epsilon-\epsilon_{\theta}\left(z_{t}, t\right)\right\|_{2}^{2}\right]    
\end{equation}
where $\epsilon$ is the ground-truth noise added to the image latent $z_{0}$ and $t$ is uniformly sampled from $\{1,...,T\}$. During the inference phase, these models progressively denoise a normally distributed variable $z_{T} \sim \mathcal{N}(0,1)$ until it reaches a clean latent $\hat{z}_{0}$. This clean latent variable can then be decoded by $\mathcal{D}$ to synthesize realistic images.

In our framework, both diffusion and denoising processes are exclusively performed in the lower half of the encoded latent, with the remaining upper half also being incorporated into denoising U-Net to provide more context. During inference, the masked input face is encoded as $z^m_0$, of which the lower half is diffused to obtain the initial $z_T$. 
During training, the ground-truth face is first encoded to $z_0$ and subsequently diffused to $z_t$ by the noise.

\subsection{Audio-driven Landmark Completion}
\label{auio-landmark}
Instead of directly conditioning the diffusion model on the audio signal, our framework first establishes the less ambiguous mapping from audio to landmarks motion of lip and jaw. 
Following~\cite{zhong2023identity}, we devise a transformer-based landmark completion module to predict the lip and jaw landmarks from the input audio. 
Specifically, we first encode the pose landmarks from the upper half of the face into a pose embedding using a 1D convolutional module. 
For facial contour information, we extract reference full-face landmarks from $N$ video frames within the input video, and encode them into $N$ reference embeddings via another 1D convolutional module.  
The mel spectrogram of the input audio is encoded into an audio embedding by a 2D convolutional module. 
These pose, reference, and audio embeddings are subsequently fed into a transformer encoder to predict the lip and jaw landmark coordinates, denoted as $\hat{C}^{\tau}_{lip} \in \mathbb{R}^{2 \times n_l}$ and $\hat{C}^\tau_{jaw}  \in \mathbb{R}^{2 \times n_j} $, respectively, where $\tau$ indicates landmarks for the $\tau$-th frame,  $n_l$ and $n_j$ are the number of landmarks used to represent the lip and jaw, respectively. To ensure temporally stable landmark prediction, we adopt the batched sequential training strategy following the common practice of previous methods~\cite{wang2022one,ma2023styletalk,wang2024styletalk++}. 
Specifically, the completion module predicts landmarks of $L$ successive frames for each video during training. 
The training objective for the landmark completion module is defined as follows:
\begin{equation}
    \mathcal{L}_{1} = \sum_{i=0}^{L-1} \left( \|\hat{C}^{\tau+i}_{lip}-C^{\tau+i}_{lip}\|_{1}+\|\hat{C}^{\tau+i}_{jaw}-C^{\tau+i}_{jaw}\|_{1} \right)
    \label{loss_L1}
\end{equation}
where $C^{\tau+i}_{lip}$ and $C^{\tau+i}_{jaw}$ are the ground-truth landmarks coordinates of lip and jaw, respectively. The predicted lip and jaw landmarks are then combined with the input pose landmarks to form the comprehensive target full-face landmarks.

However, the objective in \Cref{loss_L1} is insufficient to synchronize the lip and jaw motion with input audio due to potential inaccuracies inherent in the pre-estimated ground-truth landmarks. Therefore, we expect the predicted landmarks to be integrated into the image generation stage (\Cref{conteol_LDM}) in a differentiable manner, enabling end-to-end optimization to improve lip synchronization. 

\subsection{Inpainting Lower Half via Latent Diffusion Model}
\label{conteol_LDM}
As GANs~\cite{goodfellow2020generative} suffer from training instability and mode collapse, we resort to powerful latent diffusion models~\cite{rombach2022high} to inpaint the lower half of the face, conditioning on the completed landmarks and reference image.
For the end-to-end optimization of the whole framework and improved alignment between the reference image and the synthesized content, we introduce a novel conditioning module called TalkFormer, as demonstrated in the green section of \Cref{fig:framework}. 
TalkFormer aligns the synthesized motion with the motion represented by facial landmarks via differentiable cross-attention~\cite{vaswani2017attention}, and aligns the reference image features via an implicit warping manner implemented by another cross-attention layer. In our denoiser U-Net, TalkFormer modules exist at all scales, except the first scale which only contains residual convolution blocks. 
In the following subsections, we will detail the core components of TalkFormer and the reference appearance encoder for encoding reference facial image.

\subsubsection{TalkFormer: Align Talking Motion Differentiably} 
Previous researches~\cite{zhong2023identity,xie2021towards,zhou2020makelttalk} project the intermediate landmark representation on the image plane, forming the sketch image as generation condition in an indifferentiable manner. 
In contrast, our TalkFormer first uses a 1D-convolution embedding module to encode the target full-face landmarks from the landmark completion module into $n$ landmark embeddings $\{ e_i, i=1,2,..., n \}$, where $n$ is the number of landmarks to represent the full face.  Then, these landmark embeddings are integrated into cross-attention layers as keys and values, denoted as $K_1$ and $V_1$, respectively.  Simultaneously,  the queries $Q_1$ are extracted from the hidden features after ResNet\cite{he2016deep} blocks through an MLP layer. The output of cross-attention is computed as $Y$ according to the following equation: 
\begin{equation}    Y=\operatorname{Softmax}\left(\frac{Q_1 K_1^{\top}}{\sqrt{d_1}}\right) V_1
\end{equation}
where $d_1$ is the dimension of queries and keys. Subsequently, the results $Y$ go through a zero-initialized convolution layer and are added to the hidden features of U-Net in a residual manner. 
In this way,  the final generated face is ensured to have talking motion aligned with the motion represented by landmarks, and the diffusion model can be jointly optimized with the landmark completion module for improved lip synchronization.

\subsubsection{Reference Appearance Encoder}
To enable generalized talking face generation, a single reference face image is typically utilized as condition,  ensuring that the synthesized appearance remains consistent with the subject appearance. 
As illustrated in the pink section of~\Cref{fig:framework}, the reference face image is initially encoded to the latent space as $z_r$. To retain more fine-grained details from the reference face image, we devise an appearance encoder similar to the U-Net encoder consisting of residual convolution blocks. 
This appearance encoder converts the latent $z_r$ into multi-scale reference features symbolized as $F_{a}=\left\{F_{a}^{i} \mid i=1,2,..,I \right\}$, where $I$ represents the number of scales in U-Net. The dimensions of these features are identical to those of the hidden features in the encoder of U-Net denoiser.

\subsubsection{TalkFormer: Align Reference Appearance Features} 
To make the denoiser model aware of more appearance details from the reference image, we align the multi-scale reference appearance features in an implicit warping manner through another cross-attention layer. Specifically, we denote the hidden features after talking motion alignment as $F_h^i \in \mathbb{R}^{D \times H \times W}$, where scale $i \in \{2,3,\dots, I\}$. 
To spatially align the reference appearance features $F^i_{a}$ with the hidden features $F_h^i$, the $F^i_h$ is first transformed to the queries $Q_2 \in \mathbb{R}^{HW\times d_2}$  through an MLP layer while the $F^i_{a}$ are projected to the keys $K_2 \in \mathbb{R}^{HW \times d_2}$ and values $V_2 \in \mathbb{R}^{HW \times D}$, where $d_2$  is the dimension of the keys. Then, the correlation matrix between  $F^i_{a}$ and $F^i_h$ is computed as follows:
\begin{equation}
    S=\operatorname{Softmax}\left(\frac{Q_2 K_2^{\top}}{\sqrt{d_2}}\right)
\end{equation}
where $ S \in \mathbb{R}^{HW \times HW}$, and each element $s_{jk}$  of it indicates the semantic correspondence between the hidden feature in location $j$ and the reference feature in location $k$, with $j,k \in \{1,2,...,HW\}$. 
Based on this correlation matrix,  we can obtain the aligned reference features by referring to the relevant features in the reference appearance features.  Specifically, the reference appearance features $F^i_{a}$ are warped implicitly via a weighted sum of the values in $V_2$ as follows:
\begin{equation}
    \bar{F^i_{a}}=\operatorname{Reshape}(S V_2)
\end{equation}
where $\bar{F^i_{a}} \in \mathbb{R}^{D \times H \times W}$, and its semantic contents are spatially aligned with those of the hidden features $F^i_h$.
Eventually, the aligned reference appearance features $\bar{F^i_{a}}$ are passed through a zero-initialized convolution layer and added to the hidden features $F_h^i$ using a residual way. Consequently, the denoising process can better preserve the subject appearance details from reference images, facilitating high-fidelity talking face video generation.

\subsection{Optimization}
Benefit from TalkFormer, the joint optimization of the landmark completion module and the latent diffusion model can be achieved by employing the following objective function: \begin{equation}   
\mathcal{L}_{total} = \mathcal{L}_{ldm} + \lambda  \mathcal{L}_1
\end{equation} 
where $\mathcal{L}_{ldm}$ is the denoising objective defined in \Cref{loss_ldm} and $\lambda$ represents the weight assigned to the $\mathcal{L}_1$ loss term (\Cref{loss_L1}). In this way, the $\mathcal{L}_{ldm}$ loss will guide the landmark completion module to predict more accurate landmarks for better denoising, thus enhancing lip-audio synchronization.
Similar to the landmark completion module for improved temporal continuity, the latent diffusion model adopts the batched sequential training strategy~\cite{wang2022one,ma2023styletalk,wang2024styletalk++} where $L$ successive images are synthesized for each video during training.

\section{Experiments}

\begin{table*}[t]  
\caption{
Quantitative comparisons between the proposed and previous state-of-the-art methods in person-generic talking face video generation. 
Here $\uparrow$ denotes higher is better, and $\downarrow$ indicates lower is better.
}
\scriptsize
\setlength{\tabcolsep}{5pt}
\resizebox{\linewidth}{!}{%
\begin{tabular}{@{}l|ccccccccccccc@{}}
\toprule
\multirow{2}[4]{*}{Method} & \multicolumn{6}{c}{VoxCeleb}                                                                     &  & \multicolumn{6}{c}{HDTF}                                                                         \\ \cmidrule(lr){2-7} \cmidrule(l){9-14} 
                        & PSNR↑          & SSIM↑         & LPIPS↓         & FID↓           & CSIM↑         & SyncScore↑    &  & PSNR↑          & SSIM↑         & LPIPS↓         & FID↓           & CSIM↑         & SyncScore↑    \\ \midrule
Ground Truth            & -              & -             & -              & -              & -             & 8.59          &  & -              & -             & -              & -              & -             & 8.87          \\
Wav2Lip~\cite{prajwal2020lip}                 & {24.17}    & 0.81          & 0.173          & 78.20          & 0.52          & \textbf{9.33} &  & 22.51          & 0.78          & 0.232          & 87.99          & 0.60          & \textbf{9.43} \\
PC-AVS~\cite{zhou2021pose}                  & 21.84          & 0.76          & 0.132          & 53.72          & 0.37          & {8.31}    &  & 19.18          & 0.68          & 0.189          & 56.33          & 0.33          & {8.73}    \\
IP-LAP~\cite{zhong2023identity}                  & 24.17          & {0.83}    & {0.114}    & 44.02          & {0.55}    & 3.39          &  & {22.60}    & {0.83}    & {0.118}    & 34.92          & {0.67}    & 3.62          \\
PD-FGC~\cite{wang2023progressive}                  & 20.15          & 0.70          & 0.165          & 57.54          & 0.33          & 6.33          &  & 17.76          & 0.65          & 0.207          & 62.54          & 0.31          & 6.47          \\
DiffTalk~\cite{shen2023difftalk}                & 22.43          & 0.74          & 0.119          & {44.00}    & 0.51          & 1.42          &  & 22.03          & 0.74          & 0.121          & {30.16}    & 0.56          & 2.30          \\
\midrule
{Ours}           & \textbf{25.44} & \textbf{0.85} & \textbf{0.090} & \textbf{37.19} & \textbf{0.57} & 4.42          &  & \textbf{23.48} & \textbf{0.83} & \textbf{0.096} & \textbf{27.94} & \textbf{0.69} & 5.03          \\ \bottomrule
\end{tabular}%
}
\label{tab:table1}
\end{table*}

\subsection{Experimental Setups}

\subsubsection{Dataset}
We conduct experiments on two public audio-visual datasets, VoxCeleb~\cite{Nagrani17} and HDTF~\cite{zhang2021flow}. 
VoxCeleb is a collection of over 100,000 utterances from 1,251 celebrities, all extracted from videos uploaded to YouTube. 
HDTF is a high-resolution audio-visual dataset consisting of approximately 362 distinct videos, spanning over 15.8 hours, in 720P or 1080P resolutions. Compared to HDTF, the large-scale VoxCeleb dataset is a more standard benchmark commonly used in prior work.
To ensure a fair comparison, all comparison methods, including ours, are trained on the VoxCeleb dataset and evaluated using the test sets of both VoxCeleb and HDTF.

\subsubsection{Evaluation Metric}  We quantitatively evaluate all methods regarding visual quality and lip synchronization. 
Pixel-level visual quality is assessed through the Peak Signal-to-Noise Ratio (PSNR) and Structured Similarity (SSIM) \cite{wang2004image}, while feature-level visual quality is evaluated using Learned Perceptual Image Patch Similarity (LPIPS) \cite{zhang2018unreasonable} and Fréchet Inception Distance (FID) \cite{heusel2017gans}. Compared to pixel-level measurements, the feature-level measurements are more in line with human perception~\cite{zhang2018unreasonable,zhen2023human}.  Additionally, we employ the cosine similarity (CSIM) of identity vectors extracted by the ArcFace face recognition network \cite{deng2019arcface} to assess the preservation of subject identity. SyncScore~\cite{chung2016out} is commonly used by prior work to evaluate the lip-audio synchronization quality, despite some limitations.

\subsubsection{Comparison Methods}
We compare our approach against several state-of-the-art person-generic audio-driven talking face video generation methods. 
DiffTalk (\textit{CVPR'23})~\cite{shen2023difftalk} constructs a Diffusion-based framework for generalized talking face synthesis by conditioning the latent diffusion model on audio signal. 
PD-FGC (\textit{CVPR'23})~\cite{wang2023progressive} employs a progressive disentangled representation learning strategy to achieve fine-grained controllable talking face synthesis (e.g., eye, pose control). 
IP-LAP (\textit{CVPR'23})~\cite{zhong2023identity} is a two-stage landmark-based method that trains different stages separately and utilizes predicted optical flow to align the reference image with the target pose and expression.
PC-AVS (\textit{CVPR'21})~\cite{zhou2021pose} proposes a GAN-based framework to generate pose-controllable talking face videos by modularizing audio-visual representations. 
Wav2Lip (\textit{MM'20})~\cite{prajwal2020lip} utilizes a lip sync discriminator to guide the generator in generating lip-synced talking face videos.

\subsubsection{Comparison Setups}
\label{sef:comparison}
Wav2Lip~\cite{prajwal2020lip}, IP-LAP~\cite{zhong2023identity}, DiffTalk~\cite{shen2023difftalk}, and our method all generate talking face videos by inpainting the lower half of the face. 
Therefore, during the quantitative comparison, the lower half of the face in the input video is masked. Then, these methods reconstruct the masked area guided by the input audio and reference image. The original input video serves as the ground truth for metric calculation. 
We train DiffTalk~\cite{shen2023difftalk} using the official code until convergence, but it generates temporally unstable results. It relies on additional frame interpolation to smooth the results, affecting comparison fairness. Therefore, the frame interpolation post-processing was not employed for fair comparison. 
PC-AVS~\cite{zhou2021pose} utilizes a pose source video, an audio input, and a reference image to generate a talking face video. In our implementation, we substitute its pose source video with the ground-truth video. 
PD-FGC~\cite{wang2023progressive} requires a pose source video, an expression source video, an eye blink source video, an audio input, and a reference image to generate a talking face video. Our version replaces its pose source, expression source, and eye blink source videos with the ground-truth video.

\subsubsection{Implementation Details}
In our framework, the input face images are resized to 256$\times$256, and the latent space of  autoencoder~$\mathcal{D}(\mathcal{E}(\cdot))$ has a spatial dimension of 64$\times$64. 
Facial landmarks are extracted from video frames using the mediapipe tool~\cite{lugaresi2019mediapipe}. We represent the lip with $n_l=41$ landmarks, the jaw with $n_j=16$ landmarks, and the entire face with a total of $n=131$ landmarks. 
We set $N$ to 5 and the reference full-face landmarks are detected from the randomly selected frames of input videos. 
The hyper-parameter $\lambda$ is set to 10 and $L$ set to 5. 
To generate talking face videos, we employ the DDIM\cite{song2020denoising} diffusion sampler with 200 steps. We set the number of diffusion steps $T$ to 1000. The number of scales $I$ is 4, but we illustrate the case of $I=3$ in \Cref{fig:framework} for clarity. The reference face image can be any face image from the input video that reflects as many appearance details of the subject as possible. All comparison methods use the same reference face image to ensure a fair evaluation. Our implementation of the proposed method closely follows the code implementation of latent-diffusion~\cite{latent_diffusion}, while incorporating TalkFormer, Appearance Encoder, and Landmark Completion module\cite{zhong2023identity} as additional components.
The Appearance Encoder is designed based on the encoder structure of U-Net denoiser in latent-diffusion \cite{latent_diffusion}, excluding self-attention layers.

We train our framework on 2 NVIDIA A100(40GB) GPUs for 500 epochs with Adam optimizer\cite{kingma2014adam}. The batch size is 64, and the learning rate is 4e-5. 
The pre-trained autoencoder is frozen during training. 
The Landmark Completion module is jointly trained from scratch with the latent diffusion model. 
Our method focuses on inpainting the lower half of the face based on the input audio. Hence, to generate talking face videos, we first crop the face area from the input template video as network input. After obtaining the generated face, we employ a post-processing technique following the previous method\cite{zhong2023identity} to seamlessly blend it with the background, producing final talking face videos. 
For fair comparison, this post-processing technique was not employed during the quantitative and qualitative comparisons. 
The input template video has no length requirement and can be looped to match the length of the input audio.
We will release our code upon acceptance.
\subsection{Quantitative Evaluation}
\label{Quantitative_comparison}
We conduct a comprehensive quantitative comparison with state-of-the-art methods regarding visual quality and lip synchronization. The visual quality metrics are calculated solely based on the lower half of the generated face, since the generated upper-half face in Wav2Lip~\cite{prajwal2020lip}, IP-LAP~\cite{zhong2023identity}, DiffTalk~\cite{shen2023difftalk}, and ours almost inherit from the input video (i.e., ground truth). The comparison results are reported in \Cref{tab:table1}.

\subsubsection{Visual Quality}
Our method outperforms other methods in all visual quality metrics (PSNR, SSIM, LPIPS, FID, CSIM) on both VoxCeleb~\cite{Nagrani17} and HDTF \cite{zhang2021flow} datasets. 
Specifically, on the perceptual distance metric LPIPS and FID, our method significantly improves over other methods. This verifies that our method can produce high-fidelity talking face videos that align with human perception, preserving more appearance details. 
Besides, the highest CSIM score achieved by ours also indicates our method can preserve more identity information of the target subject. 
Although Wav2Lip exhibits a slight lag in terms of PSNR and SSIM metrics compared to our method, its FID and LPIPS values are approximately twice as high as ours, suggesting the presence of artifacts that is not aligned with human perception in their results. 
The performance of IP-LAP is closely comparable to ours in terms of the PSNR, SSIM, and CSIM metrics, but there remains a certain gap between IP-LAP and ours when assessed on the LPIPS, FID, and SyncScore metrics. 
While DiffTalk exhibits comparable performance in the FID metric, it still lags behind our approach when assessed on the LPIPS and CSIM metrics.

\subsubsection{Lip Synchronization}
Due to different speaking styles among individuals, accurate quantitative assessment of lip-audio synchronization remains a persistently challenging task. A common practice is to calculate the SyncScore based on the audio and visual features of SyncNet~\cite{chung2016out}. 
Wav2Lip, PC-AVS, and PD-FGC directly model the audio-visual mapping and obtain better SyncScore than ours. 
However, our approach notably excels in visual quality metrics, particularly in preserving the finer details of subject appearance, an aspect where others have room for improvement.
Wav2Lip utilizes SyncNet~\cite{chung2016out} as a discriminator during training. Hence, it achieves a very high SyncScore, even higher than that of the ground truth. Besides, PD-FGC and PC-AVS adopt audio-visual contrastive learning similar to SyncNet~\cite{chung2016out}, which contributes to a higher SyncScore but compromises visual quality. 
IP-LAP leverages facial landmarks as intermediate representations, but its lip synchronization is inferior to ours due to the isolated training of different stages. DiffTalk directly models the audio-visual mapping and generates temporally unstable videos with poor lip synchronization. We suspect its issue might stem from the mapping ambiguity magnified by the multi-step iteration of diffusion model.

\begin{figure*}[h!]  
  \centering 
  \includegraphics[width=0.95\linewidth]{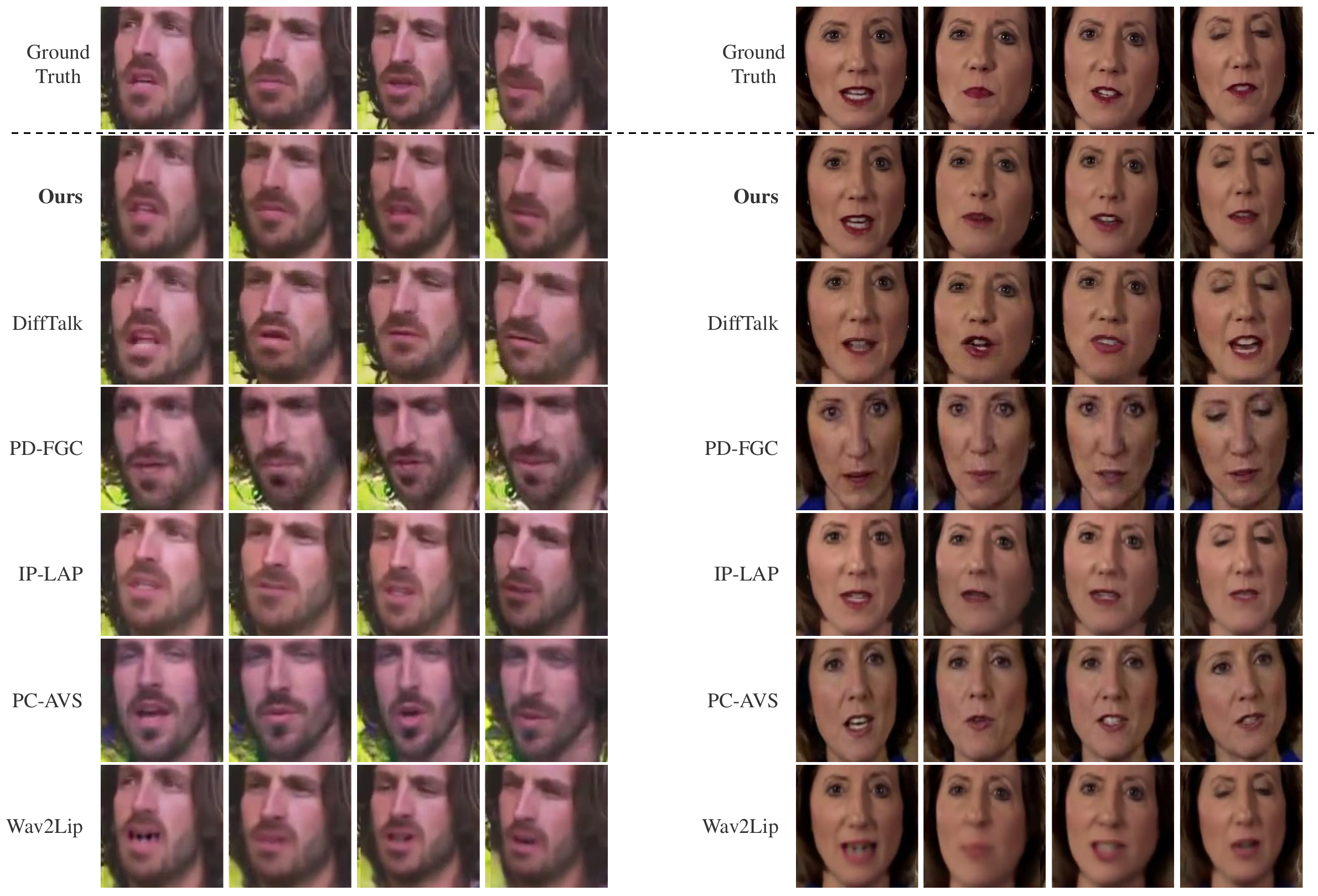}
  \caption{
  Several representative visual comparisons. The subject on the left is from the VoxCeleb~\cite{Nagrani17} dataset, while the subject on the right is from the HDTF~\cite{zhang2021flow} dataset.
  Our method achieves high fidelity of the subject appearance details with accurate lip shape. For more qualitative results, please refer to the supplementary video.}
  \label{fig:qualitative}
\end{figure*}

\subsection{Qualitative Evaluation}
\label{Qualitative_comparison}
\subsubsection{Visual Comparison}
As shown in~\Cref{fig:qualitative}, we present some representative comparison results on the HDTF~\cite{zhang2021flow} and VoxCeleb~\cite{Nagrani17} datasets. It can be observed that our results are visually closer to the ground-truth images than other methods', with more appearance details (e.g., beard, lip, teeth) preserved. It implies that our TalkFormer module could effectively align the reference image features based on semantic correlation, providing valuable features for the diffusion model. Besides, the lip shapes of ours are also closer to the ground truth. 
DiffTalk, PD-FGC, PC-AVS, and Wav2Lip directly learn the audio-visual mapping and utilize a misaligned reference image as generation condition. Therefore, their results lose some appearance details of the subject and appear blurry. 
IP-LAP generates results that are blurrier than ours and exhibit some artifacts, possibly due to the inaccurate optical flow estimation for aligning reference images. For more qualitative comparisons, please refer to the supplementary video as detailed in the following subsection. 

\subsubsection{Supplementary Video}
We have provided a short video as supplementary material. 
Please download and watch it. If there are any issues downloading the video from the review system, the same file can also be downloaded through \href{https://drive.google.com/file/d/1jZ438pPljsQJPZXViKKc7B8_7KY-aPA1}{this backup link}. The time schedule of the video is as follows:

\begin{itemize}
\item{00:00~\textasciitilde~00:08:~Video title.}
\item{00:08~\textasciitilde~00:46:~Demonstration of Ours. We demonstrate a generated result of our method where the driving audio is sourced from the text-to-speech technique, and the subject is from the HDTF\cite{zhang2021flow} dataset. We also visualize the intermediate landmarks after the landmark completion module.}
\item{00:46~\textasciitilde~01:42: Method Comparison. We present the results of all methods as well as ground-truth videos for comparison.}
\item{01:42~\textasciitilde~02:19: Ablation Study. We present the qualitative results of the ablation study, which will be further 
 analyzed in the \Cref{sec_ablation}.}
\item{02:19~\textasciitilde~03:14: More results of our method. We provide the testing results of our method for more cases.}

\end{itemize}

\subsubsection{User Study} 
For comprehensive evaluation, we conduct a user study where 16 volunteers are invited to assess the generated videos of all comparison methods.
We randomly sample 10 videos for testing, 5 from HDTF~\cite{zhang2021flow} dataset and 5 from VoxCeleb~\cite{Nagrani17} dataset.
Volunteers are asked to give their rates (0-5) for each generated video regarding image quality, lip-audio synchronization, and fidelity of appearance details. 
The videos are presented to participants in a random order and the evaluation criteria is explained in detail to the participants.
The mean opinion scores (MOS) of each method are presented in ~\Cref{tab:user_study}. Our method receives better evaluations from participants across three dimensions than other approaches.

\begin{table}[]
\centering
\scriptsize
\caption{User study results measured by Mean Opinion Scores. Scores range from 0 to 5, where higher scores denote superior performance.}
\begin{tabular}{@{}l|ccc@{}}
\toprule
Method        & Appearance Fidelity & Lip Synchronization & Image Quality \\ \midrule
Wav2Lip~\cite{prajwal2020lip}       & 2.85                & 3.67                & 2.82          \\
PC-AVS~\cite{zhou2021pose}        & 2.97                & 3.17                & 2.93          \\
IP-LAP~\cite{zhong2023identity}        & 2.88                & 2.88                & 2.79          \\
PD-FGC~\cite{wang2023progressive}        & 2.76                & 3.19                & 2.76          \\
DiffTalk~\cite{shen2023difftalk}      & 2.52                & 1.24                & 2.01          \\
\midrule
{Ours} & \textbf{4.55}       & \textbf{4.32}       & \textbf{4.54} \\ \bottomrule
\end{tabular}%
\label{tab:user_study}
\end{table}

\subsection{Ablation Study}
\label{sec_ablation}

In this section, we conduct an ablation study on the HDTF~\cite{zhang2021flow} dataset to verify the effectiveness of the proposed end-to-end framework and  TalkFormer conditioning module. The numerical results are reported in \Cref{tab:ablation}, while the qualitative results are presented in \Cref{fig:qualitative_ablate}, as well as in the supplementary video.

\begin{table}[h]
\scriptsize
\caption{Ablation study results of removing individual components of the proposed approach. 
The terms ``Ours w/o M-Align'', ``Ours w/o R-Align'' and ``Ours w/o End2End'' denote variants of our model, indicating the absence of talking motion alignment, reference appearance feature alignment, and end-to-end training, respectively.
}
\centering
\resizebox{0.95\columnwidth}{!}{%
\begin{tabular}{@{}l|cccccc@{}}
\toprule
{Variants} & PSNR↑          & SSIM↑         & LPIPS↓         & FID↓           & CSIM↑          & SyncScore↑    \\ \midrule
Ours w/o M-Align     & 20.82          & 0.75          & 0.159          & 50.28          & 0.687          & 1.12          \\
Ours w/o R-Align     & 24.28          & 0.82          & 0.114          & 43.36          & 0.790          & 1.54          \\
Ours w/o End2End     & 23.85          & 0.82          & \textbf{0.096} & 30.96          & 0.786          & 4.93          \\ \midrule
{Ours}   & \textbf{24.49} & \textbf{0.83} & 0.097          & \textbf{29.15} & \textbf{0.795} & \textbf{5.46} \\ \bottomrule
\end{tabular}%
}

\label{tab:ablation}
\end{table}

\subsubsection{Effect of End-to-End Training}
Our two-stage landmark-based method achieves the joint optimization of landmark completion module and latent diffusion model to improve lip synchronization. To validate the effectiveness of end-to-end optimization, we devise a variant where these two modules are trained separately.
Specifically, the landmark completion module is optimized using only $\mathcal{L}_1$ loss (\Cref{loss_L1}). 
In the denoiser U-Net of latent diffusion model, TalkFormer accepts the pre-estimated ground-truth landmarks as condition. The latent diffusion model is then optimized using only $\mathcal{L}_{ldm}$ loss (\Cref{loss_ldm}). 

The numerical results of this variant are reported in the ``Ours w/o End2End'' row of \Cref{tab:ablation}. In terms of visual quality metrics, this variant exhibits similar performance to our full model. However, the SyncScore of this variant drops 9.71\% compared to the full model's, verifying the effectiveness of end-to-end training in improving lip-audio synchronization. Besides, as seen in the ``Ours w/o End2End'' row of \Cref{fig:qualitative_ablate}, without the end-to-end optimization, the synthesized lip shapes are less accurate, while the visual quality remains similar to the full model's.

\subsubsection{Effect of Talking Motion Alignment in TalkFormer}
Our TalkFormer module aligns the synthesized motion with the motion represented by landmarks via cross-attention layer. 
To implement a variant without talking motion alignment, we remove the first cross-attention layer in TalkFormer that integrates landmarks embeddings as condition. Following the practice of DiffTalk~\cite{shen2023difftalk}, we redesign the landmark embedding module composed of multiple MLP layers to encode the target full-face landmarks into a single landmark embedding. This landmark embedding is added to all spatial locations of the hidden features in U-Net. 

\begin{figure}[t]  
  \centering 
  \includegraphics[width=1\columnwidth]{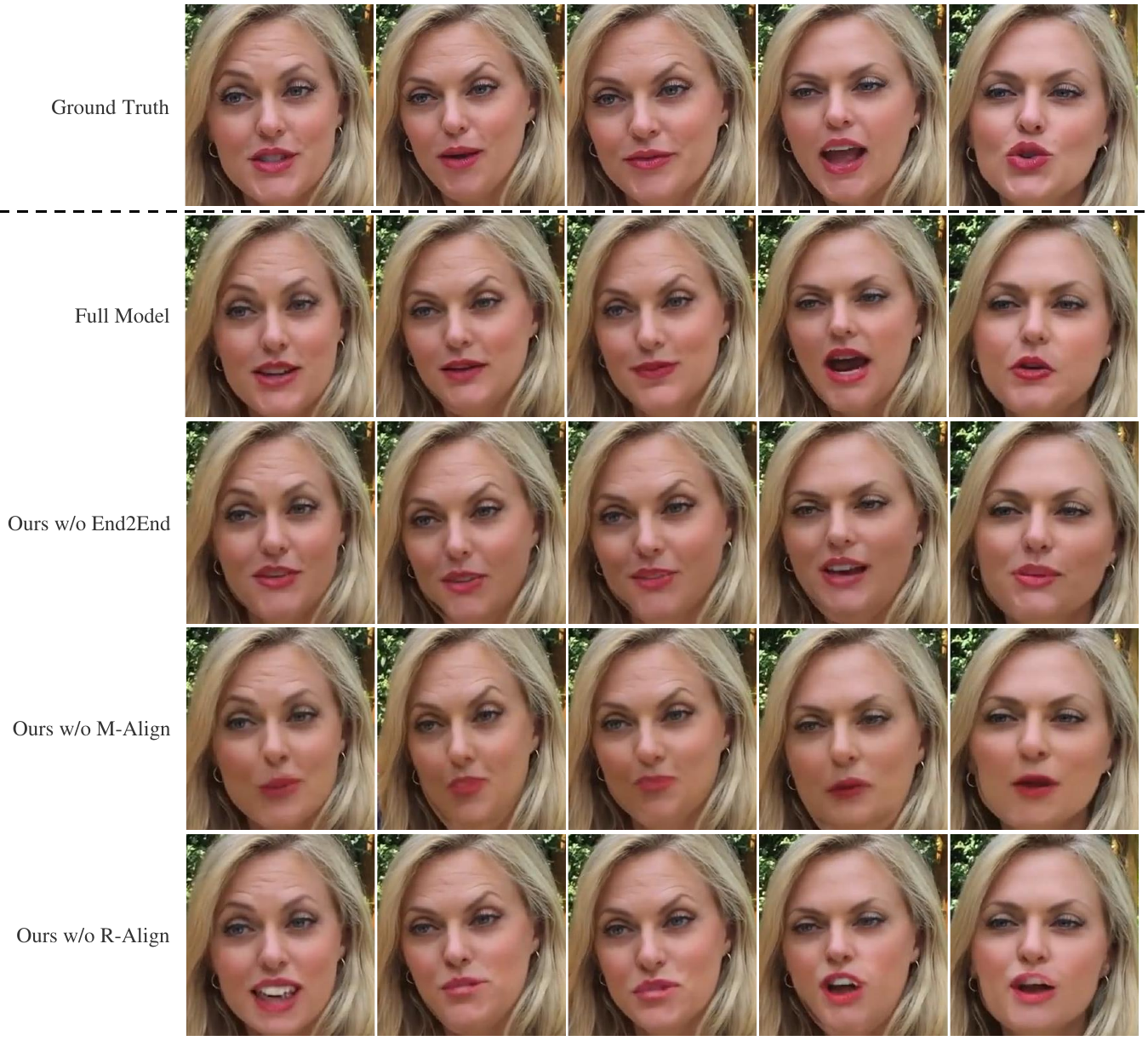}
  \caption{
  Ablation study on the effectiveness of end-to-end optimization and TalkFormer module. ``Ours w/o End2End'' represents the variant without end-to-end training. ``Ours w/o M-Align'' represents the variant without talking motion alignment in TalkFormer. ``Ours w/o R-Align'' represents the variant without reference appearance features alignment in TalkFormer.  }
  \label{fig:qualitative_ablate}
\end{figure}

The numerical results of  this variant are reported in the ``Ours w/o M-Align'' row of \Cref{tab:ablation}. It can be seen that the SyncScore drops significantly compared to the full model's. This is because the synthesized motion of the lip and jaw could not be accurately controlled through a simple addition operation. Besides, the adding operation may introduce artifacts into the generated results, deteriorating the visual quality metrics. 
As shown in the ``Ours w/o M-Align'' row of  \Cref{fig:qualitative_ablate}, in the absence of TalkFormer's talking motion alignment, the mouth shape remains predominantly closed.
Besides, influenced by the simple addition operation, the generated mouths appear blurry and exhibit some artifacts. In the first and fourth images of the ``Ours w/o M-Align'' row, the generated mouth shapes are somewhat skewed, which implies loss of the subject identity information.

\subsubsection{Effect of Reference Features Alignment in TalkFormer}
To develop a variant without reference appearance features alignment, we remove the second cross-attention layer in TalkFormer that incorporates the reference appearance features, and replace it with a self-attention layer akin to DDPM~\cite{ho2020denoising}. 
Besides, the appearance encoder is removed. Following the common practice of previous researches~\cite{shen2023difftalk,prajwal2020lip}, the reference face image is first encoded into the latent space as $z_r$. The $z_r$ is then concatenated with the noisy latent $z_t$ along the channel dimension and fed into the U-Net denoiser network. 

The numerical results of this variant are reported in the ``Ours w/o R-Align'' row of \Cref{tab:ablation}. 
It can be seen that all the visual quality metrics deteriorates compared to the full model's. Although the pixel-level metrics (PSNR, SSIM) do not change significantly, the feature-level metrics (LPIPS, FID), which are more in line with human perception, increase by a large margin. 
The potential reason is that the diffusion model can not extract meaningful features from the misaligned reference features, resulting in the loss of subject appearance details. Besides, the CSIM metric based on identity vectors might not be sensitive to subject appearance details, therefore the CSIM decreases slightly without reference features alignment in TalkFormer.
Moreover, the lip shape of the misaligned reference image might have a negative impact on the synthesized lip shape. Therefore, the SyncScore decreases without reference features alignment. 
Furthermore, as can be seen in the ``Ours w/o R-Align'' row of \Cref{fig:qualitative_ablate}, in the absence of reference appearance features alignment, the generated results lose some subject appearance details, resulting in some unrealistic contents. Besides, the lip shapes are less accurate influenced by the misaligned reference face image.

\section{Conclusion and Discussion}

In this paper, we propose a novel landmark-based framework to learn the audio-visual relationship for person-generic talking face video generation. 
Our framework utilizes facial landmarks as intermediate representations to alleviate the ambiguity of audio-visual mapping, 
while enabling end-to-end optimization to minimize error accumulation resulting from the inaccuracies of pre-estimated facial landmarks.
This accomplishment can be attributed to our innovative conditioning module TalkFormer, which aligns the synthesized talking motion with the motion represented by 
 landmarks using a differentiable cross-attention layer.
Besides, TalkFormer implicitly aligns the reference appearance features with the target motion based on semantic correlations, facilitating the preservation of more appearance details for the target subject.
Extensive experiments verify the effectiveness of our method in producing high-fidelity and lip-synced talking face videos.

\textbf{Ethical Discussion.} 
\label{sef:ethical}
Our method can generate a talking face video for any subject, requiring only a template video of the subject and a segment of audio, without the need for person-specific training. It might be misused for some illegal profits. To combat the malicious behaviors, we will watermark the generated results and are willing to contribute our synthetic videos to the deepfake detection community for enhancing their algorithms.

\bibliography{IEEEabrv,mybib}{}
\bibliographystyle{IEEEtran}

\end{document}